\documentclass[letterpaper,conference]{IEEEtran}
\usepackage[T1]{fontenc}
\IEEEoverridecommandlockouts
\usepackage{cite}
\usepackage{amsmath,amssymb,amsfonts,bm}
\usepackage{graphicx}
\usepackage{algorithm}
\usepackage{algorithmic}
\usepackage{multirow}
\usepackage{enumitem} 

\setlist{nosep} 

\begin{document}

\title{An Enhanced Proprioceptive Method for Soft Robots Integrating Bend Sensors and IMUs}

\author{Dong Heon Han, %
    Mayank Mehta, %
    Runze Zuo, %
    Zachary Wanger, %
    Daniel Bruder%
    \thanks{The authors are with the Mechanical Engineering Department at the University of Michigan, Ann Arbor, MI 48109, USA \{\tt\small dongheon, mayanknm, zuorunze, zwanger, bruderd\}@umich.edu}
}

\maketitle

\begin{abstract}
This study presents an enhanced proprioceptive method for accurate shape estimation of soft robots using only off-the-shelf sensors, ensuring cost-effectiveness and easy applicability. 
By integrating inertial measurement units (IMUs) with complementary bend sensors, IMU drift is mitigated, enabling reliable long-term proprioception.
A Kalman filter fuses segment tip orientations from both sensors in a mutually compensatory manner, improving shape estimation over single-sensor methods. 
A piecewise constant curvature model estimates the tip location from the fused orientation data and reconstructs the robot’s deformation. 
Experiments under no loading, external forces, and passive obstacle interactions during 45 minutes of continuous operation showed a root mean square error of 16.96 mm (2.91\% of total length), a 56\% reduction compared to IMU-only benchmarks.
These results demonstrate that our approach not only enables long-duration proprioception in soft robots but also maintains high accuracy and robustness across these diverse conditions.

\end{abstract}



\newcommand{\dan}[1]{{\normalsize{\textbf{({\color{blue}Dan:\ }#1)}}}}
\newcommand{\don}[1]{{\normalsize{\textbf{({\color{red}Don:\ }#1)}}}}
\newcommand{\mm}[1]{{\normalsize{\textbf{({\color{magenta}mm:\ }#1)}}}}
\newcommand{\runze}[1]{{\normalsize{\textbf{{\color{green}(runze: #1)}}}}}
\newcommand{\zach}[1]{{\normalsize{\textbf{({\color{black}Zach:\ }#1)}}}}

\newcommand{\revcomment}[2]{\textcolor{red}{#2}$^{\##1}$} 
\newcommand{\tempcomment}[2]{{#2}$^{\##1}$}    


\section{Introduction}  \label{sec:intro}
Soft robots possess intrinsic compliance and virtually infinite degrees of freedom, enabling continuous deformation \cite{rus2015soft}. 
To leverage these characteristics for safe and adaptive interaction in dynamic environments, proprioception—the ability to sense body configuration—is essential \cite{soter2018bodily}. 
However, achieving accurate proprioception remains challenging due to material flexibility and complex nonlinear deformation \cite{kramer2011soft}. 
Reliable proprioception is thus critical for enabling stable motion planning and control in soft robotic systems.  

Various transducers—resistive \cite{della2020data}, capacitive \cite{tairych2019capacitive}, inductive \cite{felt2019inductance}, magnetic \cite{mitchell2021fast}, and optical \cite{galloway2019fiber}—have been used for shape awareness in soft robots, often integrated into electronic skins \cite{shih2020electronic}. 
Yet, these require specialized fabrication, limiting accessibility. 
Recent works show that off-the-shelf sensors can offer simpler, low-cost alternatives for shape estimation.
Among such sensors, IMUs are popular for their compactness, ease of integration, and real-time tracking.  
Martin et al. \cite{martin2022proprioceptive} combined multiple IMUs with an extended PCC model \cite{webster2010design}, identifying IMU drift as a limitation. 
Our method introduces a bend-sensor reference to correct IMU yaw drift before Kalman fusion, improving long-duration stability. 
Peng et al. \cite{peng2024tendon} validated a tendon-driven manipulator with multiple IMUs but did not address drift. 
Stella et al.  \cite{stella2023soft} used a kinematics-based filter with IMU orientation data but assumed negligible twist, restricting motion to planar bending. While this simplification reduces computation, it limits robustness under torsional impact, unlike our method which handles coupled bending–twisting deformation.
Transient accelerations further amplify IMU noise and drift accumulation \cite{rong2024dynamic}, degrading accuracy in long-term operation.

\begin{figure}
    \centering
    \includegraphics[width=\linewidth]{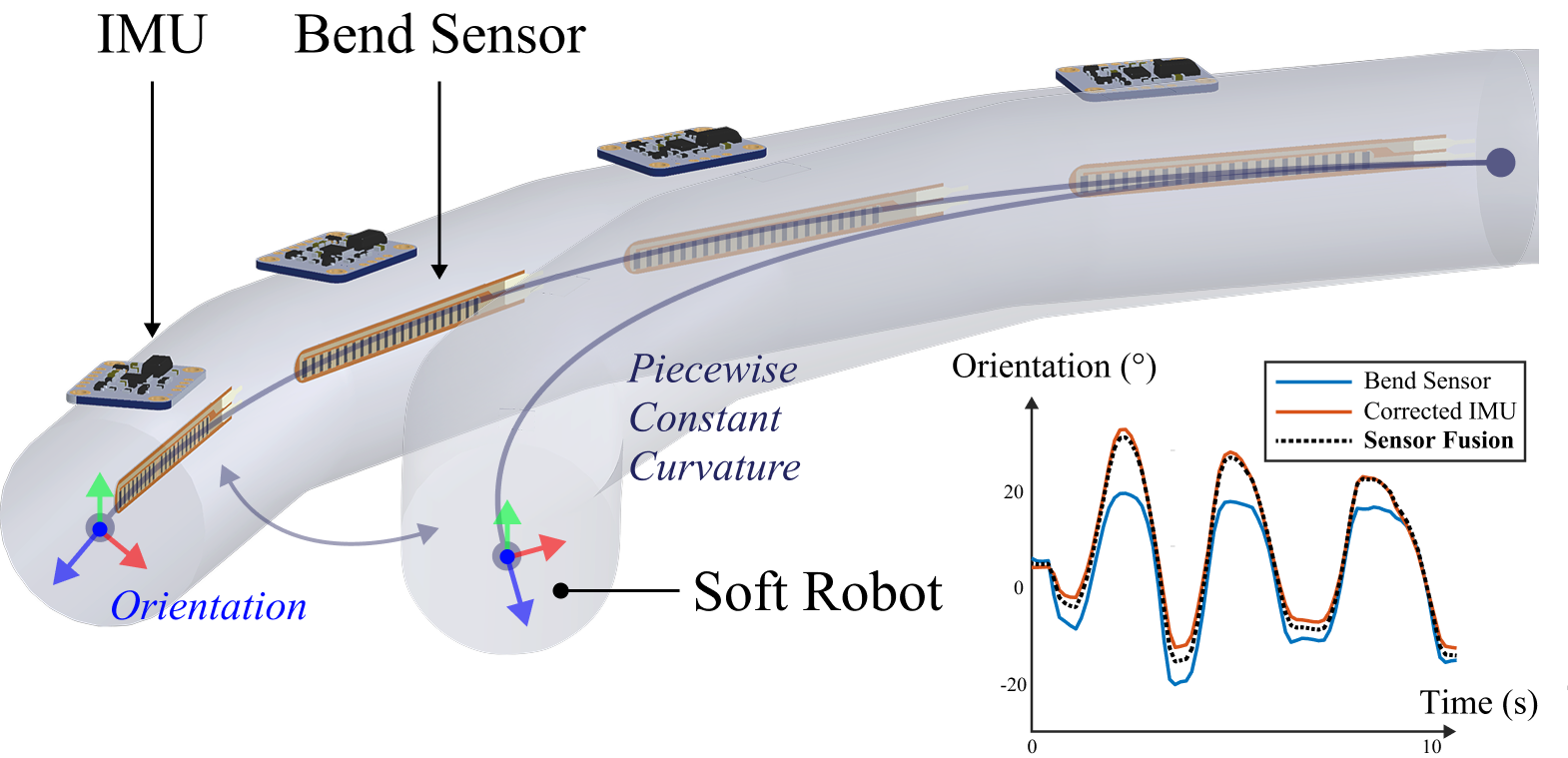}
    \caption{Overview of our proposed proprioceptive sensing method for soft robots using off-the-shelf IMUs and bend sensors. Shape estimation is based on a piecewise constant curvature model using segment-tip orientations. IMU drift is corrected using bend-sensor references before fusion, enabling 45-minute continuous operation.}
    \label{fig:fig1}
\end{figure}

To mitigate drift, previous studies fused IMUs with cameras \cite{bezawada2022shape} or draw-wire sensors \cite{stewart2022state}. 
However, camera-based fusion is affected by occlusion and lighting, and draw-wire methods require constant tension, introducing errors under compression. 
While such methods improve short-term accuracy, few demonstrate stable long-term drift correction under dynamic loads. Bend sensors offer an attractive alternative \cite{tian2023multi}, providing direct deformation measurement unaffected by lighting or tension. 
Although noisy and hysteretic, they exhibit minimal drift and insensitivity to acceleration spikes, capturing real structural changes. 
Yet, off-the-shelf bend sensors have not been systematically exploited to suppress IMU drift or sustain long-duration proprioception. 
Our approach addresses this gap by using bend sensors as low-frequency references that anchor IMU drift within a two-stage fusion pipeline for reproducible, long-term accuracy.

The Kalman filter (KF) remains a standard tool for real-time sensor fusion, combining probabilistic modeling with prediction-correction cycles \cite{maybeck1990kalman}. 
It is particularly effective for complementary fusion, integrating heterogeneous sensors \cite{sasiadek2002sensor}.  
However, its performance depends on accurate noise and covariance initialization, which is difficult in soft robots due to nonlinear material behavior \cite{wang2018toward}.  
Empirical calibration and data-driven approaches have therefore been used to model sensor deformation and improve estimation accuracy \cite{kim2021review,truby2020distributed}. 
Building on these insights, we empirically tune KF noise and covariance parameters to capture soft-robot nonlinearities, enhancing proprioceptive performance.

We propose a shape estimation framework using two off-the-shelf sensors: an IMU (BNO055, Bosch, Adafruit) and a bend sensor (2.2" Flex Sensor, Spectra Symbol), as shown in Fig.~\ref{fig:fig1}. 
Drift correction is first applied by comparing IMU and bend readings, followed by KF fusion for refined estimation. 
A PCC model maps orientation to shape, and KF parameters are optimized by minimizing RMSE against ground truth.  
We validate the approach on a 2D planar manipulator, where yaw drift—parallel to gravity—is most pronounced, offering a controlled test before generalizing to 3D environments.  

This paper’s main contributions are threefold. 
First, a compensatory drift-correction scheme anchoring IMU yaw to a calibrated bend-sensor curvature reference. 
Second, a two-stage fusion of segment-tip orientation and PCC-mapped position using linear KFs, improving accuracy without added hardware. 
Third, a fully off-the-shelf, low-cost implementation validated for 45 minutes under no-load, load, and obstacle-contact conditions, achieving a 56\% RMSE reduction versus an IMU-only PCC baseline while retaining real-time, embedded feasibility.
The Methods section details the sensing and fusion design, the Experiments section presents validation procedures, and the Discussion and Conclusion interpret results and suggest future directions.
\begin{figure}
    \centering
    \includegraphics[width=\linewidth]{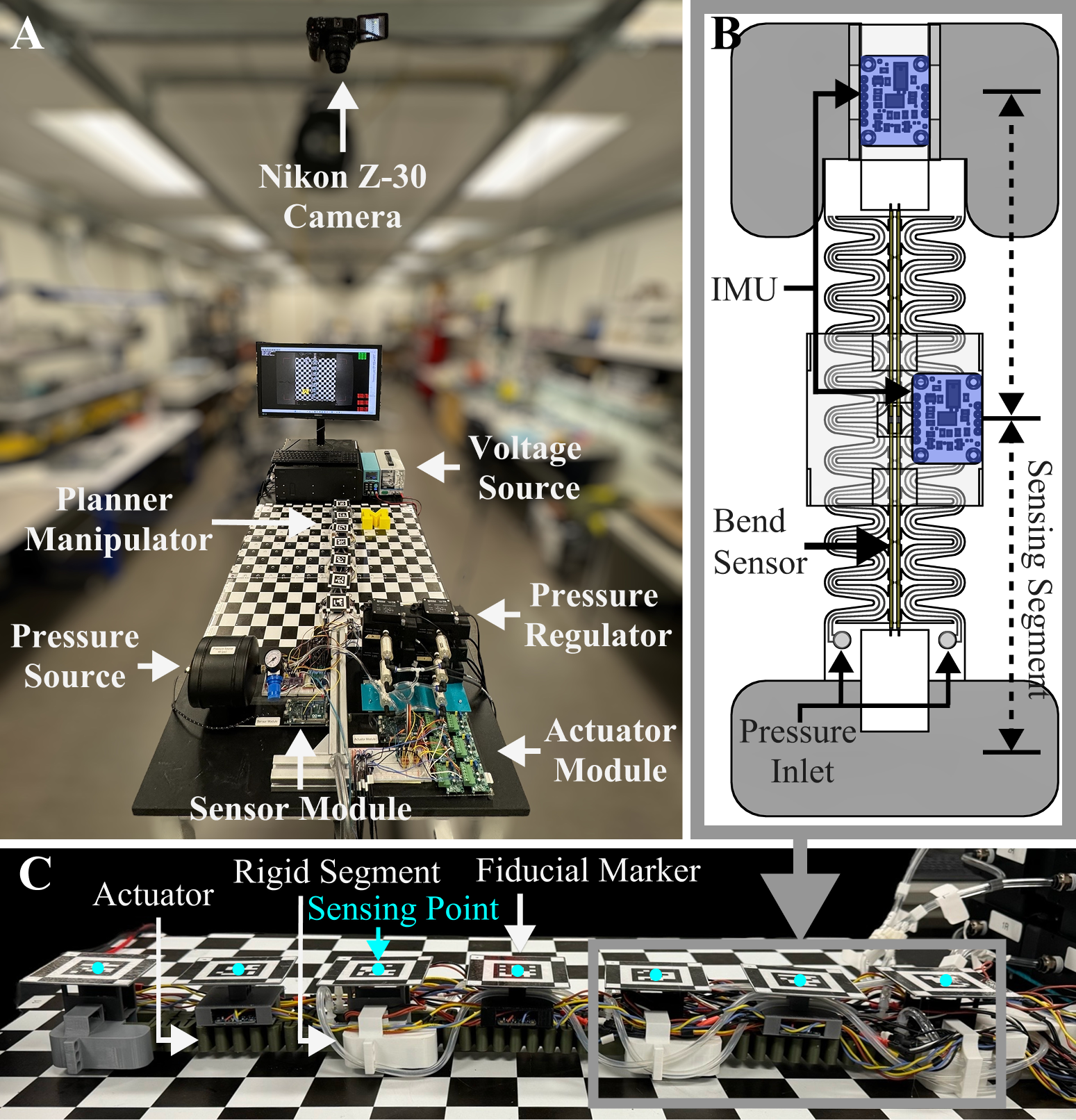}
    \caption{Experimental setup for validating the sensing method. (A) Planar robot test rig. (B) $N$th link with two actuators, four bend sensors (yellow), and two IMUs (blue) located between parallel sensors. (C) Full robot showing fiducial markers and sensing points for output comparison.}
    \label{fig:rig}
\end{figure}

\section{Methods}  \label{sec:methods}
A sensing segment is defined as the smallest unit in which multiple sensors collect data for deformation estimation. 
Each segment independently estimates its local state from these measurements. 
Proprioception is achieved using off-the-shelf IMUs and bend sensors to estimate the orientation of each segment endpoint, as illustrated in Fig.~\ref{fig:rig}. 
One end of the robot is fixed, and the structure bends laterally within a planar workspace. 
The bend sensor’s resistance varies with curvature, exhibiting a bijective relationship between endpoint orientation difference and measured resistance determined through calibration (\textbf{\textit{V–Orient Map}} in Fig.~\ref{fig:Flow}). 
To reduce noise and improve stability, two bend sensors are placed in parallel per segment, and their averaged output is used for orientation estimation, yielding both bend-sensor–mapped and IMU-derived orientations obtained from the IMU’s onboard filter. As shown in Fig.~\ref{fig:Flow}, the sensing architecture integrates these measurements to estimate each segment’s orientation and position. 
IMU drift is first corrected using the bend-sensor reference (\textbf{\textit{IMU Correction}}), and the corrected orientations are applied to the PCC model (\textbf{\textit{PCC}}) for position estimation. 
Finally, the sensor outputs are fused through two Kalman filters—\textbf{\(\textit{KF}_{\textit{orient}}\)} for orientation and \textbf{\(\textit{KF}_{\textit{coord}}\)} for position—enhancing accuracy, suppressing noise, and ensuring stable proprioceptive performance.

\begin{figure}
    \centering
    \includegraphics[width=\linewidth]{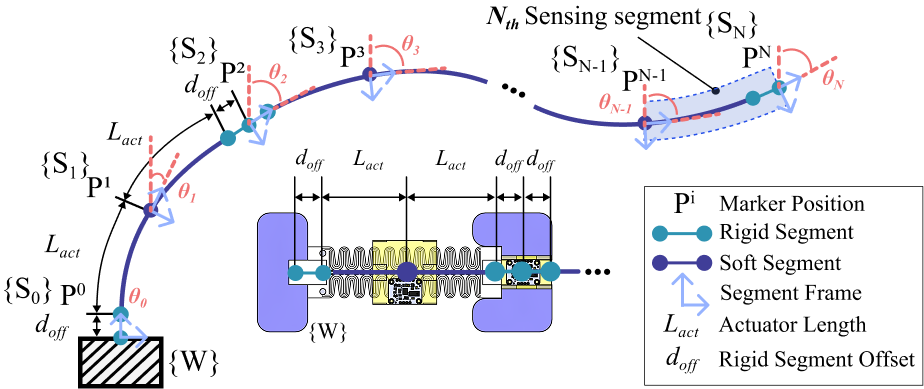}
    \caption{
    A PCC kinematic model is used to map sensor readings to an estimate of the robot's shape.
    The robot is assumed to consists of a chain of rigid and constant curvature segments.
    }
    \label{fig:Kin}
\end{figure}
\begin{figure*}
    \centering
    \includegraphics[width=\linewidth]{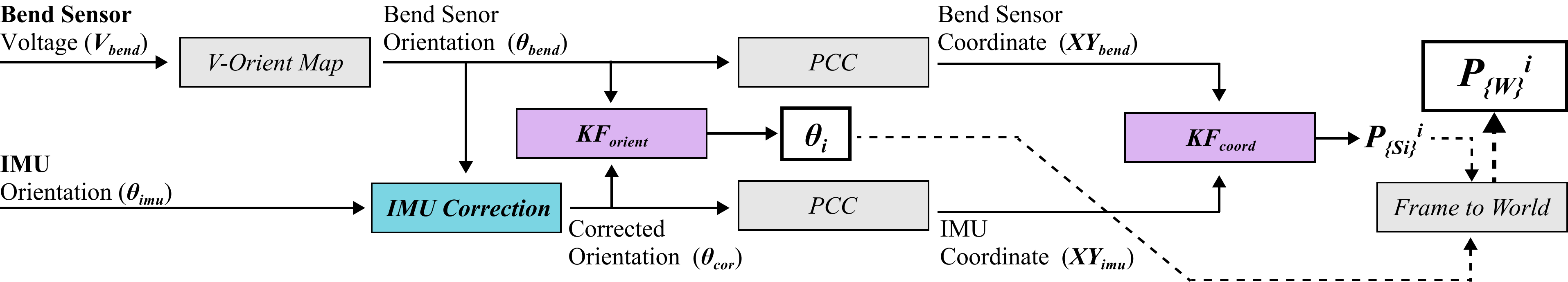}
    \caption{Flowchart showing enhanced proprioception. The process starts with signals from the bend sensor and IMU, passes through an algorithm, and results in two output data: orientation in the \({S_i}\) frame, $\theta^i$, and position in the \({S_i}\) frame, $P_{s_{i}}^i$, which are used to estimate the position of the end-effector. The orientation at each sensing point and the estimated end-effector position of each segment are then compared with the ground truth, denoted as \( GT_{\theta{\{S_i}\}}^i \) and \( GT_{P{\{W}\}}^i \), respectively, obtained from fiducial markers, and the RMSE is calculated based on the distance between the two positions.
    }
    \label{fig:Flow}
\end{figure*}
\subsection{Kinematics of Soft Manipulator}

As shown in {Figure \ref{fig:Kin}}, we choose to model the robot as a series of connected constant curvature segments. 
The robot consists of N sensing segments, each comprising a curvature with a length \(L_{\text{act}}\) and a rigid connector with an offset length \(d_{\text{off}}\). 
The \(i\)-th curvature $C^i$ is expressed in the \{S$_i$\} frame of each part as follows, where \(\theta_i\) represents the angle of the curvature, and \(\kappa_i\) represents the radius of curvature:
\begin{equation}
\mathit{C}_{\{S_i\}}^{i} = \begin{bmatrix} \kappa_i \cdot (1 - \cos(\theta_i)) \\ \kappa_i \cdot \sin(\theta_i) \end{bmatrix}
\quad \text{where} \quad \kappa_i = \frac{L_{\text{act}}}{\theta_i} \ 
\end{equation}
\(\mathit{R}_{\{S_i\}}^{i}(\theta)\) represents the rotation matrix in the \{S$_i$\} frame, and since the offset in each part can be attached either at the beginning or at the end, the \(i\)-th offset $D^i$ of the rigid connector is expressed in two cases and is represented in the \{S$_i$\} frame of each part as follows:
\begin{equation}
\mathit{D}_{\{S_i\}}^{i} = \begin{bmatrix}
d_{\text{off}} \\
0
\end{bmatrix} \cdot \mathit{R}_{\{S_i\}}^{i}(\theta) \;\;
\left(
\theta =
\begin{cases}
\theta, & \text{if } i \text{ is odd} \\
0, & \text{if } i \text{ is even}
\end{cases}
\right)
\end{equation}
Finally, the end point in the \{S$_i$\} frame is defined as \(\mathit{P}_{\{S_i\}}^{i}\) as shown below (Figure \ref{fig:Flow}):
\begin{equation}
\mathit{P}_{\{S_i\}}^{i} = \mathit{C}_{\{S_i\}}^{i} + \mathit{D}_{\{S_i\}}^{i}
\end{equation}
The six points \(P_i\) represented in the \{S$_i$\} frame can be transformed into the world frame \{W\} as follows (\textbf{$\textit{Frame to World}$} from Figure \ref{fig:Flow}):
\begin{equation}
\mathbf{P}_{\{W\}}^{i} = \left( \sum_{k=1}^{i-1} \mathbf{P}_{\{S_k\}}^{k} + \mathbf{P}_{\{S_i\}}^{i} \right) \cdot \prod_{k=1}^{i-1} \mathbf{R}_{\{S_k\}}^{k}(\theta)
\end{equation}
\subsection{IMU Drift Correction}
IMU drift accumulates over time due to the integration of noise in gyroscopic angular velocity measurements. 
In contrast, calibrated bend sensors provide stable orientation estimates since they directly measure curvature and are less affected by cumulative errors. 
Drift is mitigated using a moving average filter with a defined window size and correction threshold. 
The offset is computed as the difference between the moving averages of IMU orientation ($\boldsymbol{\theta}_{imu}$) and bend-sensor orientation ($\boldsymbol{\theta}_{bend}$) from Figure~\ref{fig:Flow}. 
When the offset exceeds the threshold, the IMU orientation is corrected by subtracting this value. 
This correction process (\textbf{\textit{IMU Correction}} in Figure~\ref{fig:Flow}) is applied to all $N$ IMUs to maintain consistent orientation accuracy. 
Although advanced onboard algorithms may further reduce long-term drift, fundamental sensor noise and nonlinearities limit purely computational solutions. 
By directly integrating bend-sensor data with IMU outputs, our method achieves real-time drift compensation, providing robust and reliable proprioception without additional algorithmic complexity.

\subsection{Sensor Fusion Using Kalman Filter}
We use linear Kalman filters for transparency and embedded suitability. The state evolves approximately linearly over small time steps. Nonlinearity is confined to the measurement mapping through PCC. We decouple orientation and mapped position to let each filter absorb different error sources.
A KF is used to fuse IMU and bend sensor data for orientation \(\theta_i\) and coordinates \((x_i, y_i)\) in the \(\{S_i\}\) frame (\textbf{\(\textit{KF}_{\textit{orient}}\)} and \textbf{\(\textit{KF}_{\textit{coord}}\)} from Figure \ref{fig:Flow}).

This process determines the orientation and position of all \(N\) segments, enabling overall robot shape estimation in the world frame, as illustrated in Figure \ref{fig:Kin}.
To model the decoupled estimation process, the system state vector is split into two separate components:
\begin{equation}
\hat{x}_{\text{orient}} = \begin{bmatrix} \theta_1 & \theta_2 & \dots & \theta_N \end{bmatrix}^T
\end{equation}
\begin{equation}
\hat{x}_{\text{coord}} = \begin{bmatrix} x_1 & y_1 & x_2 & y_2 & \dots & x_N & y_N \end{bmatrix}^T
\end{equation}
Each of these state vectors is processed independently in the KF framework, depending on whether orientation or position is being estimated. As shown in Figure \ref{fig:Flow}, estimating the shape of the robot requires two sequential steps:  
\begin{enumerate}
    \item The position of each segment’s endpoint is estimated using the PCC model, which maps endpoint positions from both the IMU and bend sensor. The resulting \(x, y\) coordinates are then fused via a Kalman Filter (\textbf{$\textit{KF}_{\textit{coord}}$}).  
    \item To reconstruct the robot’s overall shape, the relative rotation of each segment is computed. Each segment’s orientation from both sensors is fused using a Kalman Filter (\textbf{$\textit{KF}_{\textit{orient}}$}).  
\end{enumerate}
 
By combining both position and orientation estimations, the robot’s full shape can be reconstructed accurately. The general form of the Kalman filter equations remains unchanged, but the state vector used depends on the specific estimation:
\begin{equation}
\hat{x}_{k|k-1} = A \hat{x}_{k-1|k-1} + B u_k
\end{equation}
\begin{equation}
P_{k|k-1} = A P_{k-1|k-1} A^T + Q
\end{equation}
followed by a Measurement Update Step.
\begin{equation}
K_k = P_{k|k-1} H^T \left(H P_{k|k-1} H^T + R\right)^{-1}
\end{equation}
\begin{equation}
\hat{x}_{k|k} = \hat{x}_{k|k-1} + K_k \left(z_k - H \hat{x}_{k|k-1}\right)
\end{equation}
Here, \(A\) denotes the state transition matrix, \(P_k\) the state covariance matrix, \(Q\) the process noise covariance representing system uncertainties, and \(R\) the measurement noise covariance reflecting sensor noise. 
The measurement matrix \(H\) maps sensor readings to the state vector with distinct components for each sensor. 
To enhance KF performance, \(Q\), \(R\), and \(H\) are optimized via gradient descent to minimize RMSE in both orientation and \(x\)–\(y\) position estimations. 
Each parameter is iteratively updated using the RMSE gradient, allowing the filter to adapt to variations in sensor reliability and system dynamics:
\begin{equation}
p_i \leftarrow p_i - \alpha \frac{\partial \text{RMSE}}{\partial p_i},
\end{equation}
where \(p_i \in \{Q, R, H\}\) and \(\alpha\) is the learning rate. 
During optimization, \(Q\) adjusts to system uncertainty, \(R\) is tuned per sensor to reflect noise characteristics, and \(H\) refines the sensor-to-state mapping. 
This iterative process continues until the RMSE converges, ensuring accurate and robust KF estimates of both position and orientation. 
Per-cycle complexity scales linearly with the number of segments, and the computational load remains low enough for real-time execution on a microcontroller-class platform, with latency primarily limited by sensor input/output rather than filtering computation.

\section{Experiments}  \label{sec:experiments}

\subsection{Overview of the Planar Robot Test Rig}
A planar soft robot was developed to validate the proposed sensing framework, comprising control, sensing, and camera modules. 
The control module regulates pneumatic pressure across three segments of the soft arm. 
As shown in Fig.~\ref{fig:rig}, the arm consists of three 3D-printed thermoplastic elastomer (TPE) actuator modules, each used for configuration estimation. 
Each module integrates two kirigami-inspired actuators~\cite{guo2023kirigami} that bend bidirectionally under pneumatic pressure. 
Modules are connected by 3D-printed PLA joints, and metal rollers at the base enable low-friction motion. 
Pressure regulators (Enfield TR series) driven by PWM-to-analog converters modulate a 280 kPa air supply, with a microcontroller generating PWM signals for actuation.  
The sensing module integrates bend sensors and IMUs, with each actuator module forming one sensing segment. 
Bend sensors are embedded in pre-printed spinal channels, and their analog outputs are multiplexed to an Arduino Due for acquisition from twelve sensors. 
Six IMUs communicate with the same microcontroller via an I\textsuperscript{2}C expander.  
The camera module employs a ceiling-mounted Nikon Z-30 to record the workspace. 
Enhanced ArUco markers~\cite{Kedilioglu2021AruCoE} are placed at the IMU and bend-sensor locations to provide ground truth (GT) with 0.1 mm accuracy using OpenCV. 
A checkerboard background enables camera calibration via Zhang’s method~\cite{zhang2000flexible}. 
This setup provides high-fidelity validation of the soft arm’s real-time motion and configuration.

\subsection{Training and Validation}
Training data serve two purposes: (1) calibrating bend-sensor readings to match the orientations of the sensing points (Figures~\ref{fig:rig} and~\ref{fig:Kin}), and (2) identifying optimal Kalman Filter (KF) parameters using bend-sensor orientations and drift-corrected IMU data.  
The dataset comprises 40{,}000 samples collected over approximately 4{,}000 s.  
Two-thirds are obtained by sweeping the robot $\pm50^{\circ}$ for calibration, and the remaining third by generating randomized curvatures in each segment to ensure distinct actuator orientations (see accompanying video).  
All data are used exclusively for KF parameter optimization.
Validation is conducted in three scenarios to test robustness against environmental variations, obstacles, and external forces.  
In \textbf{Scenario I}, the actuators swing the robot laterally without external loads.  
\textbf{Scenario II} applies external forces using a stick, and \textbf{Scenario III} places an obstacle at the center for collision and wrapping interactions.  
Scenarios II and III evaluate the impact of external disturbances on sensing performance. 
For each scenario, the effectiveness of the PCC model and KF-based fusion is assessed, identifying the method’s strengths, limitations, and improvement areas.  
Performance is compared among four approaches: \textbf{\textit{Fusion}} (IMU + bend via KF), \textbf{\textit{Bend}} (bend-only), \textbf{\textit{IMU\(_C\)}} (IMU corrected using bend data), and \textbf{\textit{IMU\(_O\)}} (uncorrected IMU).

\section{Results}  \label{sec:results}
\begin{figure}
    \centering
    \includegraphics[width=\linewidth]{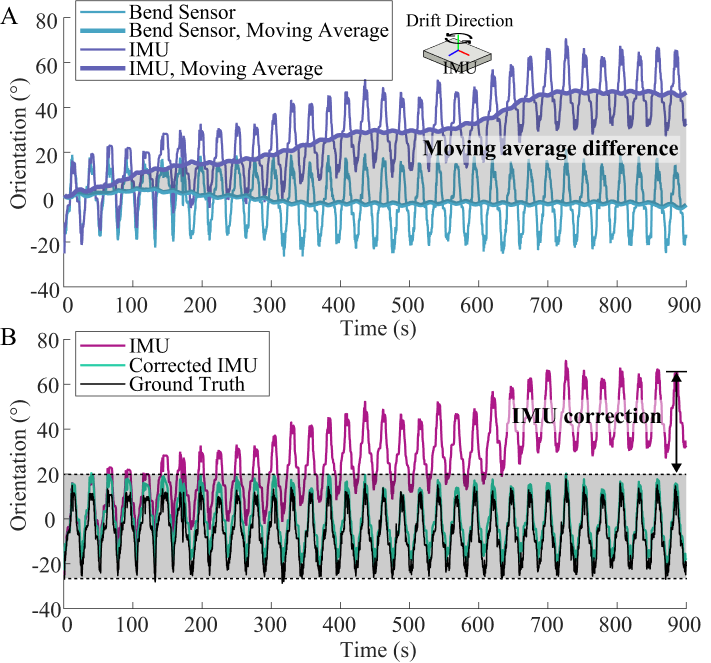}
    \caption{
    IMU drift was observed over 900 seconds (15 minutes) while periodically changing the orientation of a sensing segment between -20 and 20 degrees. (A) The figure presents the orientation trend from the bend sensor and IMU using a moving average. The IMU error accumulates over time, indicating significant drift. (B) IMU correction results are compared to raw IMU values, demonstrating that the corrected IMU closely follows the ground truth. 
    }
    \label{fig:drift}
\end{figure}

\begin{figure*}
    \centering
    \includegraphics[width=\linewidth]{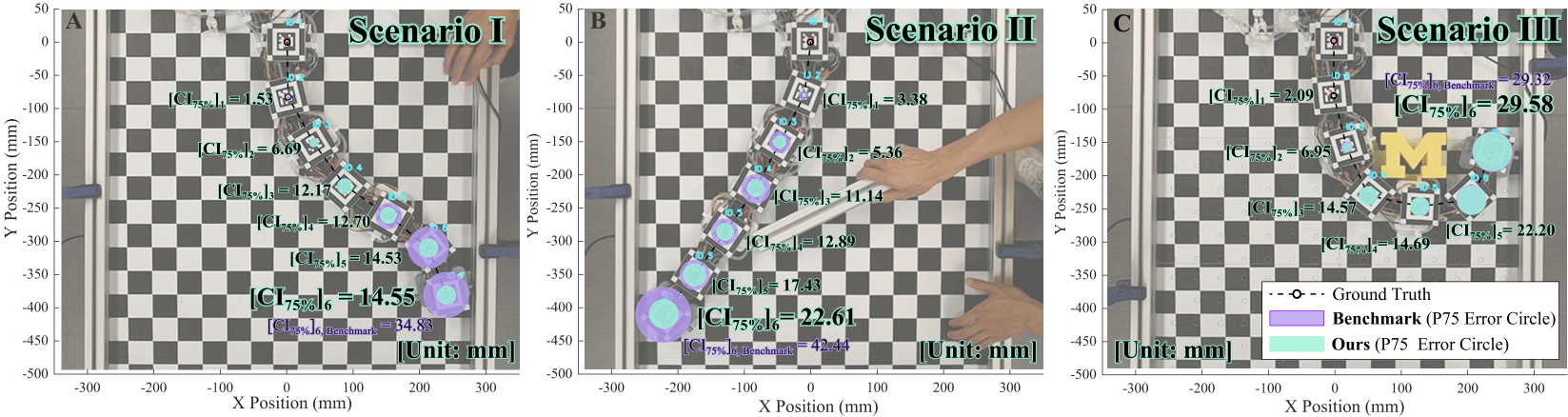}
    \caption{We compare the estimations from Scenarios I, II, and III with the actual photos taken by the camera at those times. The diameter of the circles ($\left[ CI_{75\%} \right]_i$) represents the error, defined as the Euclidean distance between the estimation from \textit{Fusion} and the ground truth, indicating the error size within the 75th percentile. This is compared to the error of \textit{IMU\(_C\)}, the benchmark ($\left[ CI_{75\%} \right]_{i, Benchmark}$).}
    \label{fig:finalestimation}
\end{figure*}

Each bend sensor is calibrated through voltage mapping and quadratic interpolation, yielding an average RMSE of 4.72° across 12 sensors. 
The mapped orientations serve as references for IMU drift correction. 
As shown in Figure~\ref{fig:drift}, this process compensates approximately 45° of drift accumulated over 900 s (15 min), preventing divergence in the IMU signal. 
The plot represents one sensing segment, with similar results observed for the remaining five. 
A moving-average window of 100 samples and a correction threshold of 0.5° are used. 
Although demonstrated during periodic motion, this correction performs robustly across diverse movements.
To assess estimation accuracy in Scenarios I–III (Figure~\ref{fig:finalestimation}, Table~\ref{tab:tb1}), the error is defined as the Euclidean distance between the estimated \(xy\) end-effector position and the ground-truth fiducial marker. 
RMSE is then computed from these errors over a 45-min experiment, validating long-term reliability. 
In Table~\ref{tab:tb1}, the first column aggregates results from \textit{S\(_I\)}, \textit{S\(_{II}\)}, and \textit{S\(_{III}\)}. 
The \textit{Fusion} method achieves the lowest RMSE of 16.96 mm (2.91 \% of total arm length), demonstrating the highest accuracy. 
By comparison, \textit{Bend} yields 24.62 mm (6.13 \%), confirming the superiority of the fusion-based approach.

Using the method \textit{IMU\(_{O}\)} yields the highest RMSE at 30.27 mm (5.20\%), representing the poorest performance. For the method \textit{IMU\(_{C}\)}, the RMSE is 17.68 mm (3.04\%), performing slightly worse than the \textit{Fusion}. 
When using method \textit{IMU\(_{0}\)}, the RMSE is 31.12 mm with an error percentage of 5.35\%, which is better than the method \textit{Bend\(_{0}\)} but still falls short compared to methods using \textit{Fusion}. 

\begin{table}[h!]
\centering
\caption{Experimental RMSE for Different Scenario Sets}
\label{tab:tb1}
\resizebox{\linewidth}{!}{
\begin{tabular}{|c|c|c|c|c|}
\hline
\multirow{2}{*}{\textbf{Methods}}                                                                                                 & \multicolumn{4}{c|}{\textbf{Scenario}}                                                                                                               \\ \cline{2-5} 
                                                                                                 & \multicolumn{1}{c|}{S$_{\textnormal{I}}$ $\cup$ S$_{\textnormal{II}}$ $\cup$ S$_{\textnormal{III}}$} & \multicolumn{1}{c|}{S$_{\textnormal{I}}$}    & \multicolumn{1}{c|}{S$_{\textnormal{II}}$}   & S$_{\textnormal{III}}$   \\ \hline
Fusion          & \multicolumn{1}{c|}{16.96 (2.91\%)*}           & \multicolumn{1}{c|}{10.97 (1.89\%)} & \multicolumn{1}{c|}{20.24 (3.48\%)} & 23.59 (4.05\%) \\ \hline
Bend              & \multicolumn{1}{c|}{24.62 (4.23\%)}           & \multicolumn{1}{c|}{19.45 (3.34\%)} & \multicolumn{1}{c|}{30.10 (5.17\%)} & 29.46 (5.06\%) \\ \hline
IMU$_{\textnormal{C}}$ & \multicolumn{1}{c|}{17.68 (3.04\%)}           & \multicolumn{1}{c|}{12.49 (2.15\%)} & \multicolumn{1}{c|}{21.88 (3.76\%)} & 22.87 (3.93\%) \\ \hline
IMU$_{\textnormal{O}}$& \multicolumn{1}{c|}{30.27 (5.20\%)}           & \multicolumn{1}{c|}{33.24 (5.71\%)} & \multicolumn{1}{c|}{36.68 (6.30\%)} & 22.75 (2.19\%) \\ \hline
\end{tabular}
}

\vspace{0.1cm} 

\parbox{\linewidth}{\raggedleft \tiny The RMSE unit is millimeters (mm).}

\vspace{0.1cm} 

\parbox{\linewidth}{\scriptsize * The percentages represent the RMSE as a fraction of the total length of the manipulator.} 
\end{table}
\textbf{Scenario~I:} \textit{Fusion} achieves the lowest RMSE of 10.97~mm (1.89\%), followed by \textit{IMU\(_C\)} at 12.49~mm (2.15\%). 
\textit{Bend} yields 19.45~mm (3.34\%), and \textit{IMU\(_O\)} produces the highest error at 33.24~mm (5.71\%).  \textbf{Scenario~II:} \textit{Fusion} again performs best with an RMSE of 20.24~mm (3.48\%), followed by \textit{IMU\(_C\)} at 21.88~mm (3.76\%). 
\textit{Bend} data results in 30.10~mm (5.17\%), while \textit{IMU\(_O\)} reaches the highest RMSE of 36.68~mm (6.30\%). \textbf{Scenario~III:} \textit{Fusion} attains 23.59~mm (4.05\%), while \textit{IMU\(_O\)} slightly outperforms it with 22.75~mm (3.92\%), representing the best performance in this case. 
\textit{IMU\(_C\)} yields 22.87~mm (3.93\%), and \textit{Bend} exhibits the largest error at 29.46~mm (5.06\%).  
Orientation errors mirror the position trends: the original IMU (\textit{IMU\(_O\)}) shows increasing yaw bias, the drift-corrected IMU (\textit{IMU\(_C\)}) removes low-frequency bias, and \textit{Fusion} further reduces orientation variance across all segments.

\section{Discussion and Conclusion}  \label{sec:conclusion}
In this study, we presented an enhanced proprioceptive method using off-the-shelf sensors to improve the versatility and reliability of soft robots. 
Our goal was to establish an accurate sensing architecture effective across diverse scenarios and long durations. 
The complementary fusion of bend sensors and IMUs was shown to significantly improve estimation accuracy. 
Validation on a 2D planar robot confirmed the method’s robustness in real time and demonstrated its potential for extension to 3D applications.

\subsection{Key Findings and Performance Evaluation}
\begin{figure}
    \centering
    \includegraphics[width=\linewidth]{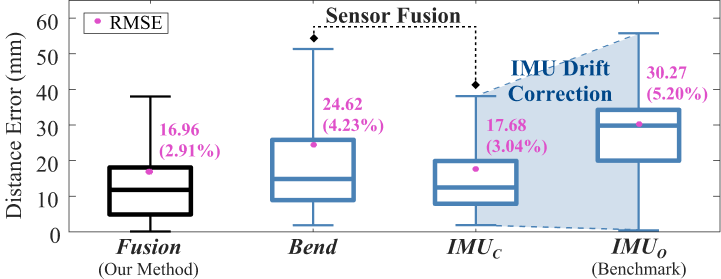}
    \caption{The figure shows the RMSE (Root Mean Square Error) of the end-effector's Euclidean distance error for Scenario I,II,III across different estimation methods. Each box plot represents the distribution of errors for a specific method, with the box spanning from the 25th percentile (Q1) to the 75th percentile (Q3), and the horizontal line within each box indicating the median (50th percentile). The stars indicate the corresponding RMSE values for each method.}\
    \label{fig:endeffectorE}
\end{figure}

IMU drift was effectively corrected using bend sensors (Fig.~\ref{fig:drift}). 
The method remained stable over 45~min and is expected to scale to longer durations. 
This drift-correction strategy generalizes to 3D configurations, offering a simple yet robust means of mitigating IMU drift. 
Bend sensors exhibited lower precision than IMUs due to resistive noise amplification, yielding higher uncertainty than \textit{IMU\(_C\)} in all scenarios. 
While IMUs are less sensitive to vibration, they remain susceptible to drift, as observed in \textit{IMU\(_O\)} under prolonged motion or external acceleration (Scenarios~I--II).
In Fig.~\ref{fig:finalestimation} and Table~\ref{tab:tb1}, \textit{IMU\(_O\)} unexpectedly outperforms \textit{Fusion} in Scenario~III. 
This anomaly arises from extreme passive deformation during obstacle contact, violating the constant-curvature assumption. 
Such conditions cause the PCC model to lose validity, degrading both bend-sensor calibration and KF tuning. 
Shorter segments or alternative curvature models may alleviate this limitation, while nonlinear or learning-based voltage--orientation mappings could improve extrapolation to large deformations. 
Outside such extremes, drift correction remains essential, and fusion of \textit{IMU\(_C\)} with \textit{Bend} consistently preserves high accuracy. 
KF gain balance depends on sensor-noise estimation: overestimating bend-sensor noise biases toward the IMU, whereas underestimation increases over-correction. 
Paired bend sensors reduce variance, and a thresholded drift anchor prevents transient instability.
Figures~\ref{fig:finalestimation}--\ref{fig:endeffectorE} summarize performance across Scenarios~I--III. 
Compared with Martin~\textit{et~al.}~\cite{martin2022proprioceptive}, which fuses multiple IMUs via PCC modeling, the proposed method achieves the primary error reduction through drift correction, with further improvement from subsequent fusion. 
Although residual outliers appear during extreme deformations, overall RMSE decreases by 56\%, yielding a 2.91\% length-normalized error sustained for 45~min without vision or custom sensors---demonstrating reliable, low-cost proprioception.

\subsection{Future Directions}
Our study validated the feasibility of sensor fusion in planar soft robots, and the next stage is extending this framework to 3D deformation.
Current commercial bend sensors are single-axis; thus, full 3D bending requires either integrating multiple uniaxial sensors as structural elements or replacing them with multi-axis curvature transducers.
This modification would be purely structural—no change to the fusion algorithm is needed, as the existing architecture naturally generalizes to three dimensions.
The system would remain real time, vision free, and robust against occlusion and lighting, ideal for embedded and wearable platforms with limited computation and power.
Computation scales linearly with segment count, and local drift correction ensures scalability for longer or more articulated robots.
Accuracy for complex configurations can be enhanced by shortening segments or adopting non-PCC curvature models.
IMU drift is continuously compensated by bend-sensor references, while long-term effects of sensor aging only affects voltage-to-orientation calibration and can be corrected by periodic recalibration.

\bibliographystyle{IEEEtran}
\bibliography{main}

\end{document}